\documentclass[letterpaper,10pt,conference]{ieeeconf}

\IEEEoverridecommandlockouts
\newlength{\topfloatinset}
\usepackage{amsmath,amssymb}
\usepackage{booktabs}
\usepackage{array}
\usepackage{makecell}
\usepackage{graphicx}
\usepackage{capt-of}
\usepackage{placeins}
\usepackage{stfloats}
\usepackage{url}
\title{\LARGE \bf
Contact-Consistent Learning of Persistent Route Memory for Blind Legged Navigation
}

\author{%
\authorblockN{Zichen Feng, Hansheng Liang$^{\dagger}$, Weifeng Huang, Bin Xu$^{*}$}
\authorblockA{School of Mechanical Engineering, Beijing Institute of Technology\\
Beijing, China\\
\texttt{bitxubin@bit.edu.cn}}}

\begin{document}
\twocolumn[{%
\renewcommand\twocolumn[1][]{#1}%
\maketitle
\begin{center}
\includegraphics[width=\textwidth]{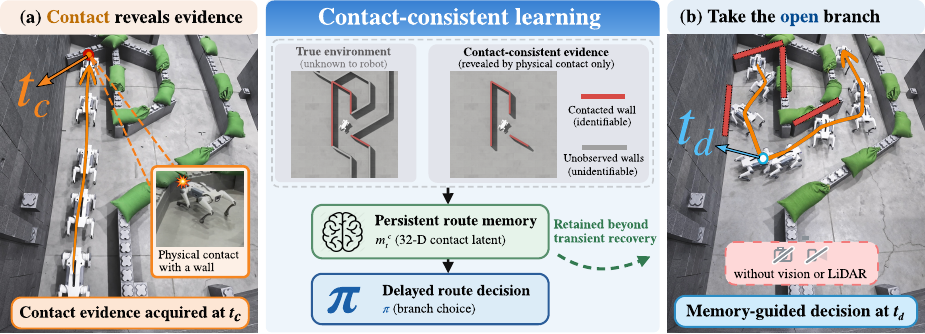}
\captionof{figure}{Collision recovery resolves the present contact. Persistent route memory determines the later branch choice. The robot contacts a blocked branch and retreats at $t_c$ (a). The contact latent persists to a later junction at $t_d$, where the open branch is selected (b). No camera or range sensor is read at deployment.}
\label{fig:recovery_memory}
\end{center}
\vspace{0.6\baselineskip}%
}]
{\renewcommand{\thefootnote}{}%
\footnotetext{\raggedright $^{\dagger}$Project leader.\quad $^{*}$Corresponding author.}}
\thispagestyle{empty}
\pagestyle{empty}

\begin{abstract}
Legged robots can recover from collisions using proprioceptive feedback, but recovery does not imply route memory: that a later decision remembers what the contact revealed. We find that widening the privileged teacher that supervises such memory from contacted wall segments to the full map makes it a stronger controller and the distilled student worse, so the supervision boundary rather than teacher strength governs what transfers. Acting on this, we present Contact-to-Latent Navigation (CTLN), whose teacher sees only wall segments revealed by physical contact. CTLN distills a 32-dimensional contact latent from proprioceptive history and exposes it as an explicit, intervenable input to the navigation actor, aligning the latent on teacher trajectories before distilling latent-guided decisions on student-induced states. On held-out delayed-decision layouts, CTLN raises multi-step cul-de-sac success from 0.57 to 0.95 and cuts blocked-branch re-entry from 0.28 to 0.03 against a capacity-matched recurrent policy. The same ordering holds on hardware: with the relative target updated onboard from proprioception alone, CTLN reaches 0.87 success on a Lite3 quadruped against 0.33 for that policy and 0.07 for reactive recovery.
\end{abstract}

\section{INTRODUCTION}

A quadruped that pushes into a wall, turns away, and keeps walking has demonstrated collision recovery; it has not necessarily learned that the contacted route is blocked. The distinction surfaces when the robot returns to a locally ambiguous junction: the body response has decayed, the current proprioceptive state resembles an earlier one, and the robot must decide whether to re-enter the route that previously stopped it.

Learning-based locomotion has enabled quadrupeds to traverse rough terrain without cameras or range sensors~\cite{hwangbo2019agile,lee2020learning,kumar2021rma,nahrendra2023dreamwaq}, and extends to other legged morphologies~\cite{liang2026moe}, which makes a frozen locomotion controller a reusable substrate for navigation. Collision-aware control detects contact from proprioceptive and inertial residuals~\cite{haddadin2017collisions,camurri2017probabilistic,lew2019contact}, recurrent policies and privileged-to-proprioceptive distillation compress temporal information into a latent state~\cite{ross2011reduction,chen2019learning,miki2022learning,ni2022recurrent}, and tactile and contact mapping writes observations into explicit maps and plans over them~\cite{suresh2021tactile,suresh2022midastouch}. These results supply the ingredients for contact-driven navigation without establishing that a legged policy preserves contact-derived evidence once the body response has passed.

The unresolved issue is a temporal and causal mismatch. Impact, posture perturbation, gait phase, and recovery dynamics are useful near the contact site and decay quickly, whereas the fact that a branch is blocked remains relevant after the robot has retreated, turned, and crossed a corridor. A generic recurrent state may mix both kinds of information~\cite{ni2022recurrent,wijmans2023emergence}, and ordinary success or collision counts do not reveal whether a later action is controlled by the content of the earlier contact. A useful definition must therefore separate immediate recovery from delayed route revision, and it must allow the remembered content to be intervened on while the current physical state is held fixed.

We define persistent route memory behaviorally. Let $t_c$ denote the end of a contact episode and $t_d>t_c$ a later decision. If two rollouts have matched current physical and navigation states at $t_d$ but different contact histories, persistent memory is present when the policy shifts its route preference in the history-consistent direction. The shift must survive a delay and controlled intervening motion, and it must be directional rather than an arbitrary action divergence. This definition turns a vague claim about hidden state into a falsifiable delayed-decision test.

The natural way to supply this evidence is to give a privileged teacher the full maze and distill it. That teacher controls better and the student it produces is worse, because part of what the teacher conditions on never reaches the student's sensors. We therefore blind the teacher as well, revealing obstacle geometry only after a physical contact event, so its labels stay tied to what joint torques, acceleration, and command-response mismatch can show. The contact latent is the representation this produces; persistent route memory is the behavior it supports. The student first learns to reproduce that latent, then to use it on states its own actions produce.

We make three contributions. First, we identify the supervision boundary that makes route memory learnable, and show that widening it degrades the student it produces. Second, we present CTLN, whose recurrent proprioceptive encoder produces a 32-dimensional contact latent exposed as an explicit and intervenable actor input. Third, we give a falsifiable behavioral criterion for delayed route memory and the matched-state protocol that tests it.

\section{RELATED WORK}

\subsection{Proprioceptive and Blind Locomotion}
Blind locomotion methods infer terrain and body dynamics from joint states, inertial measurements, actions, and temporal context~\cite{hwangbo2019agile,lee2020learning,kumar2021rma,nahrendra2023dreamwaq,miki2022learning}. Their objective is stable locomotion under missing exteroception, and the recovered latent describes the terrain, contact state, and actuator response relevant to the current step~\cite{ji2022concurrent}. We keep that sensing boundary and ask what the representation retains once the disturbance is over.

\subsection{Collision-Aware Navigation and Recurrent Policies}
Collision-aware control detects contact from proprioceptive residuals~\cite{haddadin2017collisions} and proprioceptive signals also support terrain assessment during navigation~\cite{elnoor2024pronav}, with contact treated as a usable measurement rather than only a fault~\cite{camurri2017probabilistic,lew2019contact}. Recurrent policies encode short histories under partial observability~\cite{ni2022recurrent}, learned agents maintain explicit or emergent spatial memory for navigation~\cite{chaplot2020learning,kwon2023rnrmap}, and teacher--student methods transfer privileged information into deployable observations~\cite{ross2011reduction,chen2019learning}. Learned legged navigation couples such policies to a frozen locomotion controller and traverses clutter at speed, but with exteroception~\cite{agarwal2022legged,cheng2024extreme,roth2024viplanner,hoeller2024anymal,ren2024topnav}; resilient local navigation additionally reconstructs environment information in latent space from degraded exteroceptive input and reacts to body collisions end-to-end~\cite{zhang2024resilient}. Closest to our question, blind navigation agents trained only from collision and egomotion signals develop map-like structure in their recurrent memory~\cite{wijmans2023emergence}; that study uses an idealized point agent in a simulated indoor benchmark and reads the memory with decoding probes, whereas we study a dynamically locomoting quadruped whose contacts are physical impacts and test the memory by intervening on it at the policy interface. Immediate recovery and short-horizon inference do not provide a behavioral test for persistent route memory.

\subsection{Tactile and Contact Mapping}
Contact mapping systems accumulate contact observations in explicit spatial representations for localization, shape estimation, exploration, or planning~\cite{suresh2021tactile,suresh2022midastouch,kossas2024whisker}, and related occupancy and terrain mapping pipelines navigate unknown environments once observations are registered against an odometry estimate~\cite{duong2020autonomous,miki2022elevation}. CTLN differs in where the memory lives and how it is learned: its contact latent is distilled from a privileged, contact-consistent map that is discarded after training.

\section{PROBLEM DEFINITION}
\label{sec:problem}

\begin{table}[!htb]
\centering
\caption{Interfaces, fixed configuration, and budget. Deployed parameters cover the student, the odometry estimator, and the frozen controller.}
\label{tab:config}
\scriptsize
\begin{tabular}{lc@{\hskip 1.4em}lc}
\toprule
Component & Value & Component & Value\\
\midrule
Student frame & 62 & Student head & 128\\
Navigation obs. & 12 & TBPTT truncation & 64 steps\\
Navigation action & 3 & Map crop & $6\times6$ m\\
Teacher map & $2\times20\times20$ & Contact force gate & $5$ N\\
Map encoder out & 32 & Contact distance gate & $0.08$ m\\
GRU layers / hidden & $2/256$ & Control step & $0.02$ s\\
Deployed parameters & 1.73 M & Teacher env. steps & 492 M\\
Student / estimator & 866 k / 676 k & Student env. steps & 717 M\\
\bottomrule
\end{tabular}
\end{table}

\begin{figure*}[!t]
\vspace*{\topfloatinset}%
\centering
\includegraphics[width=\textwidth]{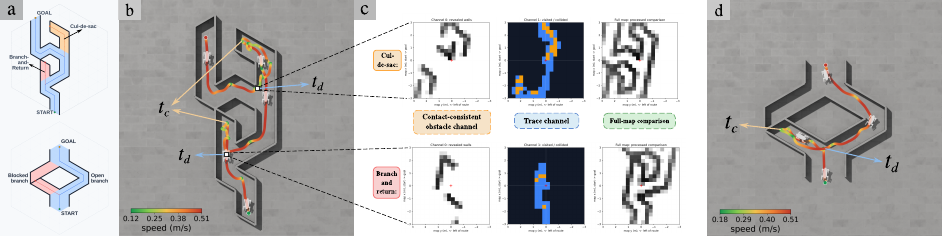}
\caption{Task layouts, teacher input, and CTLN rollouts. (a) Branch-and-Return with a Multi-Step Cul-de-Sac (top) and the matched-state paired layout (bottom). (b) A CTLN episode on the cul-de-sac layout, coloured by body speed and marked with the revealing contact at $t_c$ and the scored decision at $t_d$. (c) The teacher map $M_t$ at the scored junction of each task. The contact-consistent obstacle channel carries only wall segments the robot physically touched and the trace channel only ground it has visited or collided with, so both stay inside the sensing boundary; the full-map comparison adds every wall the robot never met and is the ablation of Sec.~\ref{sec:teacher}. (d) One illustrative CTLN trajectory on the paired layout, not a scored pair.}
\label{fig:tasks}
\end{figure*}

\begin{figure*}[t]
\vspace*{\topfloatinset}%
\centering
\includegraphics[width=0.80\textwidth]{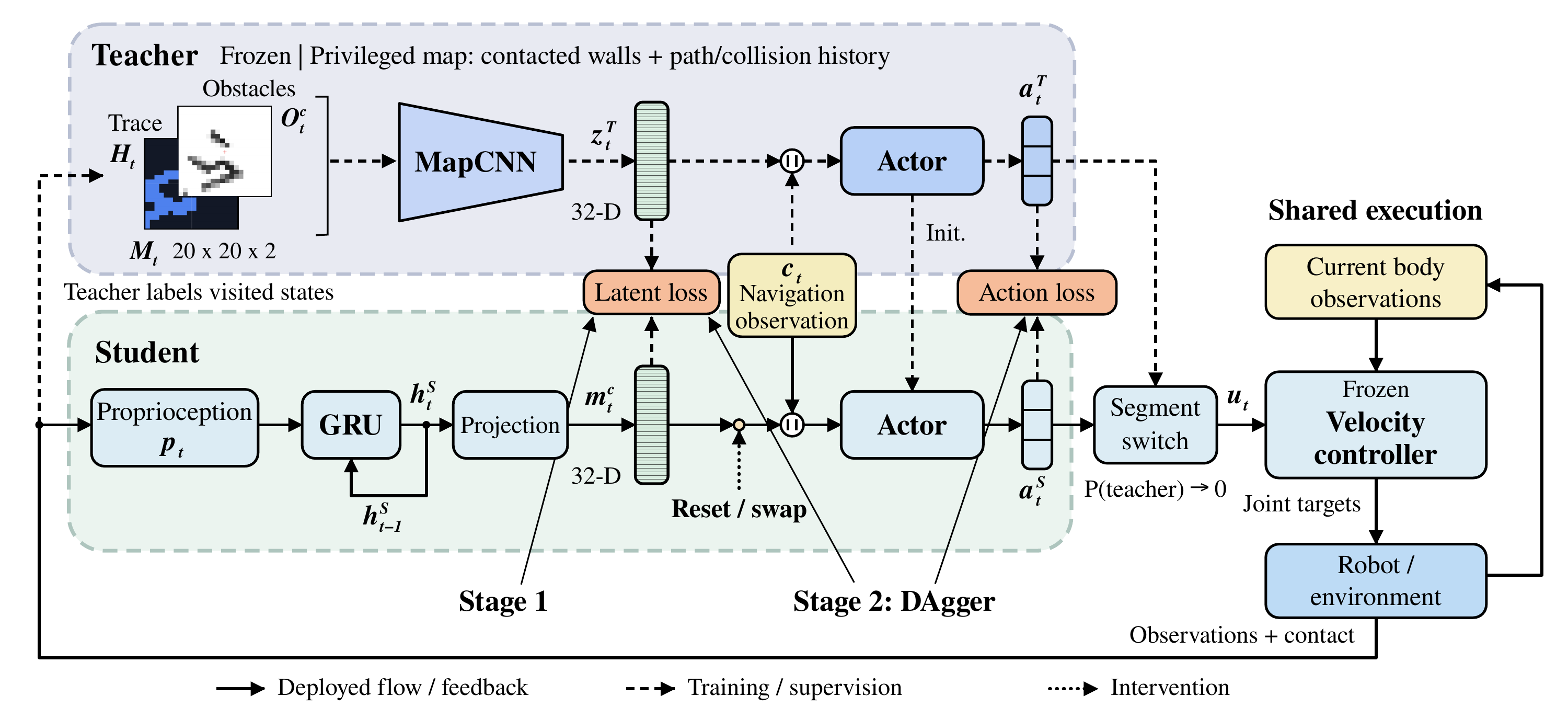}
\caption{CTLN information flow. Privileged contact-consistent map supervision is used only during training; the deployed actor receives the current navigation observation and an intervenable contact latent.}
\label{fig:overview}
\end{figure*}

We consider a quadruped navigating a procedurally generated maze toward a relative target. The navigation command is
\begin{equation}
a_t^N=(v_{x,t},v_{y,t},\omega_{z,t}),
\end{equation}
which is converted to joint targets by a frozen low-level velocity controller~\cite{rudin2021learning}. Deployment observations contain joint positions and velocities, applied torques, the previous low-level action, body angular velocity, projected gravity, body linear acceleration, the current velocity command, and a relative yaw pair. The student does not read camera, depth, LiDAR, obstacle geometry, or a pre-built obstacle map. Let $p_t$ be the 62-dimensional proprioceptive frame, $c_t$ the 12-dimensional current navigation observation, and $M_t$ the privileged contact map, which has two $20\times20$ channels: a contact-consistent obstacle channel and a trajectory/collision trace channel. The teacher receives $[c_t,M_t]$; the student receives $p_t$ and $c_t$ but not $M_t$. Table~\ref{tab:config} lists these interfaces and the fixed configuration, and Fig.~\ref{fig:overview} summarizes the resulting information flow.

The route decision is defined at a junction rather than at the collision itself. As shown in Fig.~\ref{fig:tasks}a, a trial first directs the robot into a branch that terminates at a physical wall. After contact, the robot must detach, retreat, and return to the junction, where the second decision is scored against the branch that remains open. This construction prevents a controller from receiving credit for simply reversing the current command. Multi-Step Cul-de-Sac inserts a longer detour before the escape decision, so the contact-to-decision interval spans additional time, distance, and turns.

\section{CONTACT-TO-LATENT NAVIGATION}

\subsection{Contact-Consistent Teacher Representation}

The teacher is deliberately denied the complete maze layout. Its local map is $M_t=[O_t^c,H_t]$, shown in Fig.~\ref{fig:tasks}c at the scored junction of each task: $O_t^c$ holds only the wall segments the body actually struck, $H_t$ the ground already crossed and collided with, and the full map adds the walls that were never met. Let $\mathcal{K}_t$ be the set of revealed segments, $F_t^h$ the strongest horizontal contact force, $d_t$ the closest link-to-wall distance, and $k_t^*$ the segment selected by that contact. With $\bigvee$ the elementwise disjunction of binary grids, the contact channel and its update are
\begin{equation}
\begin{aligned}
O_t^c&=\bigvee_{k\in\mathcal{K}_t}\operatorname{Rasterize}(k),
\qquad
\mathcal{K}_{t+1}=\mathcal{K}_t\cup\{k_t^*:\chi_t=1\},\\
\chi_t&=\mathbf{1}[F_t^h\geq F_{\min}]\,\mathbf{1}[d_t\leq d_{\max}],\\
&\quad\ F_{\min}=5\,\mathrm{N},\qquad d_{\max}=0.08\,\mathrm{m}.
\end{aligned}
\end{equation}
The trace channel $H_t$ is updated from the swept footprint, with visits and collision traces assigned distinct values and stronger collision marks retained. Both channels reset at episode boundaries and goal-switch teleports. The crop is a yaw-aligned $6\times6$ m body-frame window sampled on a $20\times20$ grid, so the teacher summarizes local revealed evidence rather than the persistent global metric grid maintained by elevation and occupancy mapping pipelines~\cite{duong2020autonomous,miki2022elevation}. The gate defines the information boundary, which is the gap between the first two channels of Fig.~\ref{fig:tasks}c and the third: a wall that is visible in the simulator but untouched cannot alter $z_t^T$, and a repeated touch adds no segment. Sec.~\ref{sec:teacher} removes the gate and measures what the restriction buys.

The frozen map encoder and teacher actor are
\begin{equation}
z_t^T=E_T(M_t),\qquad a_t^T=\pi_T([c_t,z_t^T]),
\end{equation}
where $E_T$ is the convolutional map encoder. Changing $M_t$ therefore changes only evidence that could have been acquired by contact. The teacher is trained with PPO and then frozen, with the configuration in Table~\ref{tab:teacher}.

\begin{table}[!htb]
\centering
\caption{Teacher training. Unlisted PPO terms are library defaults.}
\label{tab:teacher}
\scriptsize
\begin{tabular}{lc@{\hskip 0.9em}lc}
\toprule
Item & Value & Item & Value\\
\midrule
Algorithm & PPO, Adam & Iterations & 10{,}000\\
Learning rate & $10^{-3}$ & Batch / iter. & 49{,}152\\
Target KL & 0.01 & Episode length & 100 s\\
Entropy coef. & $2\times10^{-3}$ & Actor / critic & 512--256--128\\
Clip / $\gamma$ / $\lambda$ & 0.2 / 0.99 / 0.95 & Teacher params & 0.997 M\\
\midrule
\multicolumn{4}{l}{Command limit: $\pm$0.5, 0.3 m/s, 1.2 rad/s}\\
\multicolumn{4}{l}{Rewards: goal $+10$, safe direction $+3$, velocity direction $+2$, stuck $-5$}\\
\multicolumn{4}{l}{Safe direction is scored against the contact-consistent map, not the true layout.}\\
\bottomrule
\end{tabular}
\end{table}

\subsection{Proprioceptive Contact Latent}

The recurrent input is the 62-entry proprioceptive frame of Sec.~\ref{sec:problem}. The goal is excluded from it. The current navigation observation is fused only after the contact latent has been estimated. A two-layer gated recurrent unit~\cite{cho2014learning} (hidden width 256) followed by a 128-unit head produces
\begin{equation}
\begin{aligned}
h_t^S&=\operatorname{GRU}(p_t,h_{t-1}^S),\\
m_t^c&=g(h_t^S)\in\mathbb{R}^{32},\qquad
\hat a_t=\pi_S([c_t,m_t^c]).
\end{aligned}
\end{equation}
Here $h_t^S$ is the computational state of the estimator and $m_t^c$ the contact latent consumed by the actor. The hidden state persists across control steps, resets at episode and goal-switch boundaries, and is detached every 64 steps for truncated backpropagation without being erased. Truncation bounds credit assignment rather than the forward horizon: the state crosses truncation boundaries unchanged, and every step supplies its own target $z_t^T$, so retention is learned as a per-step update instead of through a gradient path spanning the full contact-to-decision delay. The actor never receives $h_t^S$: only $m_t^c$ crosses the actor interface.

Routing only $m_t^c$ to the actor makes the intervention target coincide with the variable trained to match $z_t^T$. A swap applied to the full recurrent state would also change locomotion stabilization, phase, and contact recovery. The encoder still sees the complete proprioceptive frame, so torque spikes, command-response mismatch, and inertial transients are integrated before the latent is exposed. The latent can therefore be swapped between matched observations while the physical state is held fixed, or reset to a no-contact history. Sec.~\ref{sec:causal} scores the resulting conditions.

\subsection{Two-Stage Closed-Loop Distillation}

Stage 1 answers what should be remembered. The environment is driven by the frozen teacher. At each step, the teacher produces $z_t^T$ from the contact-consistent map and the student predicts $\hat z_t$. We optimize a visibility-balanced regression objective
\begin{equation}
\mathcal{L}_{\mathrm{latent}}=
\frac{\sum_t w_t\lVert\hat z_t-z_t^T\rVert_2^2}{\sum_t w_t},
\end{equation}
where the latent regression uses weight $1$ when the contact-consistent obstacle channel is nonempty and weight $0.2$ otherwise. This visibility balancing prevents the large number of uninformative empty views from dominating the representation target while retaining empty views needed to learn the no-contact state.

Stage 2 answers how the representation is used under the student's own consequences. Following interactive imitation learning, which corrects teacher--student covariate shift~\cite{ross2011reduction}, the executed action at fixed control segments is sampled as
\begin{equation}
\begin{aligned}
u_t&=b_k a_t^T+(1-b_k)\,\operatorname{stopgrad}(\hat a_t),\\
b_k&\sim\operatorname{Bernoulli}(\beta_k),
\end{aligned}
\end{equation}
where $b_k$ is held for a multi-step segment and $\beta_k$ is annealed from teacher control to zero. The teacher labels the actual visited state regardless of $b_k$. The joint objective is
\begin{equation}
\mathcal{L}_{\mathrm{joint}}=
\lambda_z\mathcal{L}_{\mathrm{latent}}+
\lambda_a\operatorname{SmoothL1}\left(
\frac{\hat a_t-a_t^T}{s_a}\right),
\end{equation}
where $s_a$ is the per-action normalization, with $\lambda_z=0.25$ and $\lambda_a=1$. The segmented mixture exposes the student to contacts, retreats, stalls, revisits, and route corrections that are absent from a purely offline teacher rollout, and the teacher recomputes its labels from the contact map those consequences actually generate. The latent term stays active in Stage 2, so action fitting cannot repurpose the bottleneck as an unconstrained shortcut. Teacher control is sampled once per 100-step segment and annealed linearly to zero over 50\% of Stage 2.

\subsection{Deployment Goal Update}

The main memory experiments use a ground-truth relative target to isolate route-memory behavior. For onboard closure, a separate student-driven proprioceptive velocity estimator predicts body-frame velocity from a history of proprioceptive frames, following legged state estimators that fuse leg kinematics, contact, and inertial measurements~\cite{ji2022concurrent,lin2021legged,buchanan2021learning}. Integrating the estimated planar velocity with IMU yaw gives an estimated position and hence a relative target update:
\begin{equation}
\hat v_t=g_\psi(p_{t-H_o+1:t}),\qquad
\hat x_{t+1}=\hat x_t+R_z(\psi_t)\hat v_{t,xy}\Delta t.
\end{equation}
This proprioceptive velocity-integrated odometry (PIVO) closes the target loop onboard and is the condition under which every hardware comparison in Sec.~\ref{sec:deploy} is run.

\subsection{Sim-to-Real Transfer}

Transfer uses domain randomization over link friction ($[0.3,1.0]$), link mass and inertia ($\times[0.85,1.15]$), actuator gains, and AR(1) IMU noise. External pushes are disabled because a push is proprioceptively indistinguishable from a wall contact and would corrupt the evidence the latent is trained to encode. The contact gate is a simulator privilege: $F_t^h$ and $d_t$ are physics-engine quantities, and no contact detector runs onboard. What transfers is the student's regression from the 62-dimensional proprioceptive frame.

\section{EXPERIMENTS}

\begin{table*}[!t]
\vspace*{\topfloatinset}%
\centering
\caption{Delayed route-memory navigation on held-out layouts, over 5 seeds $\times$ 20 episodes. Parentheses give 95\% $t_4$ intervals across seed means, or Clopper-Pearson over the 100 pooled episodes where seed variance is zero. SPL is scored on goal-reaching. Ablations follow the same protocol and are scored on Branch Success and Re-entry only.}
\label{tab:main}
\small
\resizebox{0.92\textwidth}{!}{%
\begin{tabular}{llccccc}
\toprule
Method & Memory & Branch Success $\uparrow$ & Cul-de-sac Success $\uparrow$ & Re-entry $\downarrow$ & SPL $\uparrow$ & Repeat Contact $\downarrow$\\
\midrule
Blind + goal & None & 0.08 \scriptsize(0.00--0.17) & 0.00 \scriptsize(0.00--0.04) & 0.07 \scriptsize(0.01--0.13) & 0.00 \scriptsize(0.00--0.04) & 94.8 \scriptsize(72.3--117.3)\\
Reactive recovery & Reflex & 0.05 \scriptsize(0.00--0.13) & 0.00 \scriptsize(0.00--0.04) & 0.14 \scriptsize(0.03--0.25) & 0.00 \scriptsize(0.00--0.04) & 61.8 \scriptsize(41.2--82.4)\\
Short-window & Local history & 0.86 \scriptsize(0.73--0.99) & 0.20 \scriptsize(0.07--0.33) & 0.56 \scriptsize(0.37--0.75) & 0.09 \scriptsize(0.03--0.15) & 43.4 \scriptsize(34.7--52.1)\\
Matched GRU & Unstructured recurrent & 0.93 \scriptsize(0.86--1.00) & 0.57 \scriptsize(0.40--0.74) & 0.28 \scriptsize(0.12--0.44) & 0.28 \scriptsize(0.20--0.36) & 25.8 \scriptsize(14.6--37.0)\\
\textbf{CTLN} & \textbf{Contact latent} & \textbf{1.00} \scriptsize\textbf{(0.96--1.00)} & \textbf{0.95} \scriptsize\textbf{(0.89--1.00)} & \textbf{0.03} \scriptsize\textbf{(0.00--0.09)} & \textbf{0.50} \scriptsize\textbf{(0.46--0.53)} & \textbf{12.9} \scriptsize\textbf{(6.8--19.1)}\\
\midrule
\multicolumn{7}{l}{\textit{CTLN ablations}}\\
\quad w/ full-map teacher & Contact latent & 0.81 \scriptsize(0.70--0.92) & -- & 0.36 \scriptsize(0.28--0.44) & -- & --\\
\quad w/ contact channel only & Contact latent & 0.84 \scriptsize(0.64--1.00) & -- & 0.50 \scriptsize(0.35--0.65) & -- & --\\
\quad w/ trace channel only & Trace latent & 0.28 \scriptsize(0.12--0.44) & -- & 0.29 \scriptsize(0.21--0.37) & -- & --\\
\quad w/ offline teacher rollouts & Contact latent & 0.61 \scriptsize(0.52--0.70) & -- & 0.50 \scriptsize(0.44--0.56) & -- & --\\
\bottomrule
\end{tabular}
}
\end{table*}

\subsection{Experimental Setup}

We evaluate CTLN in Isaac Lab~\cite{mittal2023orbit} (formerly Orbit), a learning framework built on NVIDIA Isaac Sim that supersedes Isaac Gym~\cite{makoviychuk2021isaac}, on a Deeprobotics Lite3 quadruped driven by the shared frozen low-level velocity controller~\cite{rudin2021learning}. Layouts are procedurally generated mazes of full-height walls inside a 7.2 m-circumradius hexagonal room of 1.2 m triangular cells, leaving 0.889 m corridors, with start and goal 5.0--7.4 m apart in a straight line and 6.0--20.4 m along the route. Training, validation, and test layouts are separated by generator seed and topology, and the test set is never used to select intervention thresholds. Simulation results average 5 training seeds over 20 held-out episodes per seed, paired-history tests use 5 seeds over 20 matched pairs, and hardware results use three session batches of ten trials ($n=30$). Episodes are capped at 100 s, and an emergency stop is triggered by a joint-limit violation or an IMU-detected unsafe attitude.

We report the quantities listed in Tables~\ref{tab:main},~\ref{tab:causal}, and~\ref{tab:real}, including SPL~\cite{anderson2018evaluation}, computed per episode or per matched pair. Table~\ref{tab:main} varies one factor. Every row shares the low-level controller, the sensing boundary, the target interface, the action limits, the layout distribution, the evaluation protocol, and, where learned, the 62-dimensional frame, recurrent width, and optimizer budget. The rows differ only in what persists between the revealing contact and the scored decision: nothing, a reflex, a bounded window, an unstructured recurrent state, and an explicit contact latent. Blind locomotion~\cite{hwangbo2019agile,lee2020learning,kumar2021rma,nahrendra2023dreamwaq} with a goal command tests whether strong low-level control is sufficient. Reactive recovery responds to the current collision from a 0.5 s proprioceptive history and retains nothing beyond it, and short-window resets the recurrent context at a fixed horizon. Matched GRU carries the same class of unstructured recurrent memory studied in~\cite{wijmans2023emergence} into our sensing boundary. It shares the teacher, the two-stage schedule, and the recurrent capacity, and differs only in that the actor reads the recurrent state directly with $\lambda_z=0$, so the CTLN row isolates the contact latent bottleneck and its per-step supervision.

\subsection{Delayed Route-Memory Navigation}

Table~\ref{tab:main} reports both tasks on held-out layouts. Branch Success records selection of the open branch on the second visit to the junction, and Re-entry records selection of the previously blocked branch.

The first two rows show that the task is solved by neither locomotion nor reflexes. Blind locomotion with a goal reaches 0.08 Branch Success, reactive recovery 0.05, and neither escapes a multi-step cul-de-sac in any episode, while both accumulate the largest repeated-contact counts in the table (94.8 and 61.8). That count is not normalized by episode duration, so these two rows are also inflated by the time they spend unsolved. Their low Re-entry values, 0.07 and 0.14, measure how rarely they reach a second decision at all rather than any route memory. Short-window history is enough to pass the first junction at 0.86 and not enough to carry the evidence through a detour: 0.20 cul-de-sac success with 0.56 re-entry. The capacity-matched GRU closes much of that gap at 0.93 and 0.57 using the same recurrent width, observations, and training budget, which places the remaining difference in how the evidence is represented rather than in whether recurrence is present. CTLN selects the open branch in every held-out Branch-and-Return episode, escapes 0.95 of the cul-de-sac layouts, and re-enters a known blocked branch in 0.03 of them. Fig.~\ref{fig:tasks}b and~\ref{fig:tasks}d show CTLN clearing each layout.

SPL and Repeat Contact show that this is not bought with a longer or more abrasive route: CTLN roughly doubles the matched GRU's SPL, 0.50 against 0.28, and halves its repeated contacts, 12.9 against 25.8. The interventions that follow ask whether that capability is produced by the contact content the policy is supposed to store.

\begin{figure*}[!t]
\vspace*{\topfloatinset}%
\centering
\includegraphics[width=0.92\textwidth]{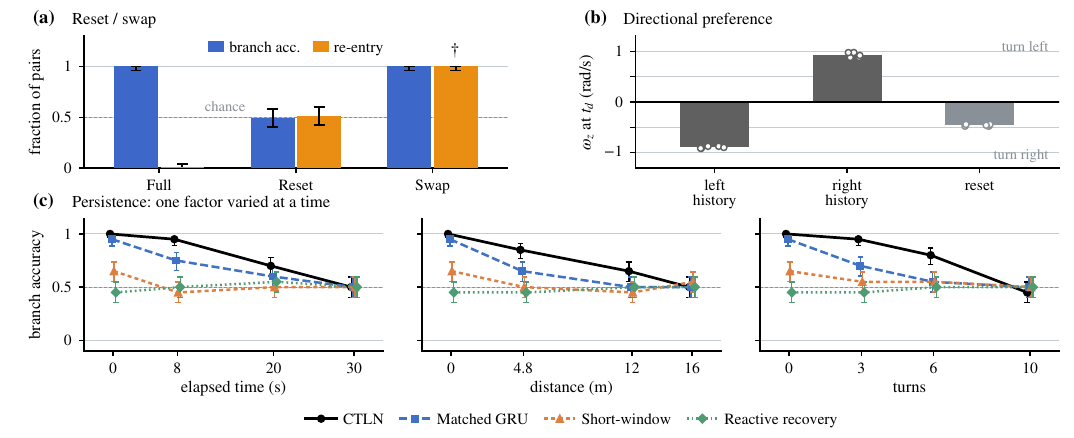}
\caption{Causal content and persistence of the contact latent. Panels (a) and (b) intervene on a single CTLN checkpoint; $\dagger$ marks the swap bar, whose two metrics are scored against different layouts and are therefore not complementary (Sec.~\ref{sec:causal}). Panel (c) compares trained policies, with 95\% intervals and a small horizontal offset for legibility.}
\label{fig:causal_persistence}
\end{figure*}

\subsection{Causal Content and Persistence}
\label{sec:causal}

A pair of episodes (Fig.~\ref{fig:tasks}a, bottom) is admissible if, at $t_d$, the robot pose, velocity, target condition, and local navigation observation match within pre-defined tolerances while the histories differ in which branch was physically contacted. Pairs are balanced over blocked side, initial orientation, target direction, and trial order. Both episodes of a pair enter the decision from the same position and attitude, with the body state restored before either policy is queried. Paired branch accuracy is the fraction of pairs whose selected branch matches the condition's reference history: the robot's own experienced history for the unmodified policy, or the injected history under the counterfactual swap. Every intervention is applied at the actor interface, so the locomotion state is untouched.

\begin{table}[!htb]
\centering
\caption{Paired-history interventions, over 5 seeds $\times$ 20 pairs. Intervals follow Table~\ref{tab:main}, with pooled pairs in place of pooled episodes.}
\label{tab:causal}
\small
\resizebox{\columnwidth}{!}{%
\begin{tabular}{lccc}
\toprule
Intervention / Variant & Paired Branch Acc. $\uparrow$ & Paired Re-entry $\downarrow$ & Success $\uparrow$\\
\midrule
Full CTLN & 1.00 \scriptsize(0.96--1.00) & 0.00 \scriptsize(0.00--0.04) & 1.00 \scriptsize(0.96--1.00)\\
Contact-latent reset & 0.49 \scriptsize(0.40--0.58) & 0.51 \scriptsize(0.42--0.60) & 1.00 \scriptsize(0.96--1.00)\\
Counterfactual swap & 1.00 \scriptsize(0.96--1.00) & 1.00 \scriptsize(0.96--1.00) & 0.60 \scriptsize(0.47--0.73)\\
Contact-only & 0.72 \scriptsize(0.51--0.93) & 0.28 \scriptsize(0.07--0.49) & 0.77 \scriptsize(0.64--0.90)\\
Trace-only & 0.49 \scriptsize(0.38--0.60) & 0.51 \scriptsize(0.40--0.62) & 0.41 \scriptsize(0.23--0.59)\\
Short-window & 0.62 \scriptsize(0.46--0.78) & 0.38 \scriptsize(0.22--0.54) & 0.95 \scriptsize(0.84--1.00)\\
\bottomrule
\end{tabular}
}
\end{table}

The question is whether the encoder builds the blocked side out of raw proprioception and how long it survives there. Full CTLN selects the history-consistent branch in every pair. Resetting the latent while leaving the current observation intact drops paired branch accuracy to 0.49, at chance: nothing in the current state identifies the blocked side, so the encoder assembled it from earlier torques. Injecting the paired history confirms the actor reads that channel, restoring branch accuracy to 1.00 and driving Paired Re-entry to 1.00, because the injected history marks the opposite side as blocked and the robot walks into the branch blocked in its own layout, with success falling to 0.60. The 0.60 that remains comes from episodes where the resulting collision writes a new segment into the latent and the robot escapes on the corrected evidence. In this row alone the two columns are not complementary, because Branch Acc. is scored against the injected history and Re-entry against the robot's own layout. The commanded yaw rate at $t_d$ carries the same signature: left-history pairs command $-0.89$ rad/s and right-history pairs $+0.92$ rad/s, with the sign consistent across every pair in all 5 seeds. Reset collapses this to a fixed $-0.46$ rad/s that no longer tracks the history, which is why the paired choice falls to chance.

Under the paired protocol the blocking fact again comes from the contact channel, with the trace channel disambiguating which revealed segment applies (Sec.~\ref{sec:teacher}). The short-window row bounds the duration of the effect at 0.62 paired accuracy. Its 0.95 success comes from eventual escape, not from a correct first choice.

Figure~\ref{fig:causal_persistence} sweeps the contact-to-decision interval one factor at a time until accuracy reaches chance. CTLN holds 0.70 branch accuracy at 20 s, 0.65 at 12 m, and 0.80 after six turns, returning to chance at 30 s, 16 m, and ten turns. The matched GRU falls faster on every axis and is already there at 12 m. Short-window sits at chance from the shortest inserted interval onward, and reactive recovery throughout. The decay is gradual rather than a cliff at a context boundary, and the sweep locates the horizon.

\subsection{A Stronger Teacher Is a Worse Supervisor}
\label{sec:teacher}

The lower block of Table~\ref{tab:main} reads four CTLN variants on the same layouts, metrics, and protocol as the main comparison, so each variant sits on the same axis as the CTLN row. Widening the teacher from contact-consistent segments to the full map drops branch success from 1.00 to 0.81 and raises re-entry from 0.03 to 0.36, against a teacher that itself reaches 1.00 success on every layout. The same question applies to which channel the teacher shows. Contact evidence carries most of the signal at 0.84 branch success, the trace channel alone collapses to 0.28, and only the joint channel reaches 1.00.

\subsection{Fitting the Teacher Is Not Acting Where the Student Goes}

Offline teacher rollouts reach 0.61 branch success with 0.50 re-entry while fitting the teacher better than any other configuration, with an action error of 0.007 on teacher-visited states against 0.038 for CTLN. Fitting the teacher where the teacher goes is not the same as acting well where the student goes: the decisive states are the contacts, retreats, stalls, and revisits that the student's own mistakes produce, and those never appear in a teacher rollout.

\begin{figure*}[!b]
\centering
\includegraphics[width=0.88\textwidth]{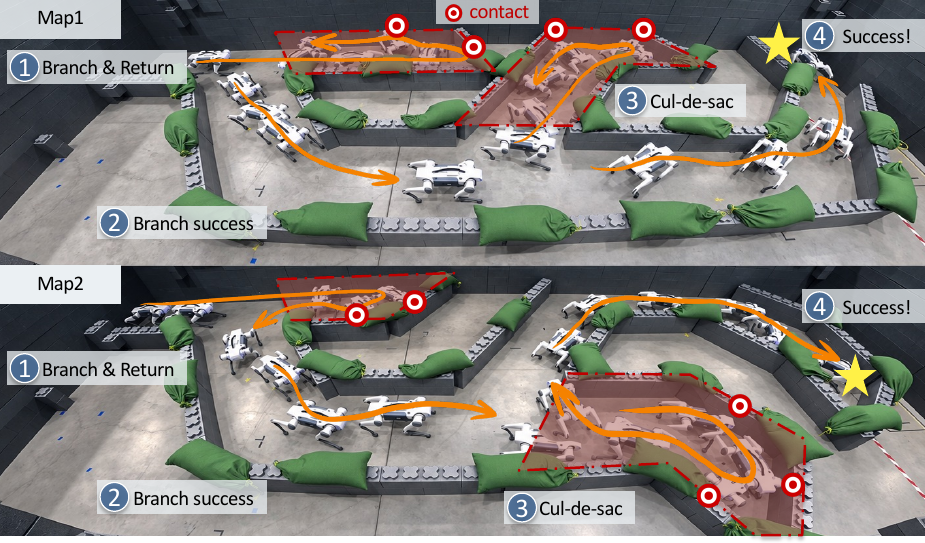}
\caption{Fully onboard CTLN deployment on two real layouts. Each row composites successive frames of one trial with the executed route overlaid: the robot contacts the blocked branch and retreats (1), takes the open branch on its return (2), clears the multi-step cul-de-sac (3), and reaches the goal (4). The accompanying video shows complete uncut trials on both layouts.}
\label{fig:real_boundary}
\end{figure*}

\subsection{Quadruped Deployment}
\label{sec:deploy}

Hardware tests what simulation cannot: with no contact detector onboard, the encoder must infer contact events from proprioception alone. Every hardware comparison runs with the relative target closed onboard by PIVO, so no method receives external localization. Starting pose, blocked side, initial orientation, and trial order are balanced across conditions. Success requires clearing both a real Branch-and-Return layout and a real Multi-Step Cul-de-Sac layout. A supplementary CTLN-GT-goal condition replaces the onboard target with an externally supplied one. It sits outside the ranking and serves only to separate odometry error from latent error.

\begin{table}[!htb]
\centering
\caption{Real-robot delayed decisions on both layouts, over three sessions of ten trials. Success and Re-entry use Wilson intervals over the 30 trials. Count columns are per-trial means with $t_2$ intervals over sessions, and a zero count uses a rule-of-three bound. Ranked rows share a fully onboard target update. CTLN-GT-goal is a supplementary diagnostic with an external target.}
\label{tab:real}
\scriptsize
\resizebox{\columnwidth}{!}{%
\begin{tabular}{lcccc}
\toprule
Method & Success $\uparrow$ & Re-entry $\downarrow$ & Collision Count $\downarrow$ & Emergency Stop $\downarrow$\\
\midrule
Reactive & 0.07 \scriptsize(0.02--0.21) & 0.77 \scriptsize(0.59--0.88) & 76.0 \scriptsize(41.2--110.8) & 3.67 \scriptsize(2.23--5.10)\\
Matched GRU & 0.33 \scriptsize(0.19--0.51) & 0.47 \scriptsize(0.30--0.64) & 45.0 \scriptsize(13.3--76.7) & 1.67 \scriptsize(0.00--4.54)\\
\textbf{CTLN} & \textbf{0.87} \scriptsize\textbf{(0.70--0.95)} & \textbf{0.17} \scriptsize\textbf{(0.07--0.34)} & \textbf{24.0} \scriptsize\textbf{(17.4--30.6)} & \textbf{0.00} \scriptsize\textbf{(0.00--0.10)}\\
\midrule
\textit{CTLN-GT-goal} & \textit{0.93} \scriptsize\textit{(0.79--0.98)} & \textit{0.00} \scriptsize\textit{(0.00--0.11)} & \textit{13.7} \scriptsize\textit{(9.9--17.5)} & \textit{0.00} \scriptsize\textit{(0.00--0.10)}\\
\bottomrule
\end{tabular}
}
\end{table}

Table~\ref{tab:real} reproduces the simulation ordering on hardware. Reactive recovery succeeds in 0.07 of trials, re-enters the blocked branch in 0.77, and averages 3.67 emergency stops. The matched GRU reaches 0.33 success with 0.47 re-entry. CTLN reaches 0.87 with 0.17 re-entry and no emergency stop, and cuts the overt collision count from 76 and 45 to 24. Figure~\ref{fig:real_boundary} shows this configuration on both layouts: the robot contacts the blocked branch, retreats, crosses to the second junction, clears the multi-step cul-de-sac, and reaches the goal from proprioception alone.

The supplementary row locates CTLN's residual hardware failures. Supplying the target externally raises success to 0.93 and removes re-entry entirely, so what remains at 0.87 is odometry error rather than latent error. PIVO integrates a proprioceptive velocity estimate with IMU yaw, and its position error accumulates with distance travelled and with every collision-induced slip. The failures occur late in trials where repeated retreat attempts and wall contacts have already displaced the estimated pose: the latent still encodes which side was blocked, but the junction it is anchored to has drifted, so the policy applies correct evidence at the wrong place.

At the 10 m scale used here the terminal position error stays within 0.5 m, below the 0.889 m corridor width; at 100 m it would not, so localization rather than latent persistence caps route scale.

\section{DISCUSSION AND CONCLUSION}

Because the actor consumes the contact latent and nothing else besides the goal, the evaluation can replace it. Clearing the latent returns the route choice to chance while the current observation is untouched, and the remaining signal survives 20 s, 12 m, and six turns of intervening motion. That is stronger than showing recurrence helps and narrower than claiming a semantic world model~\cite{hafner2025mastering}. Privileged distillation is usually tuned by making the teacher stronger. These ablations invert that prescription: what the teacher knows matters less than whether the student can identify it from its own sensors, so a teacher that is better at the task can be the worse supervisor. Contact is a clean place to test this because the boundary is a physical event: a wall is either touched or not, so the identifiable subset of the map is decidable instead of estimated.

These results characterize what a proprioception-only policy retains between a contact and a later decision. They do not measure how that retention compares with route memory built from exteroception, which would require a different sensing boundary. The deployed system is anchored to a proprioceptive pose estimate, which caps route scale and produces every residual hardware failure we observe.

\FloatBarrier

\end{document}